\documentclass[runningheads]{llncs}

\usepackage{eccv}

\usepackage{eccvabbrv}

\usepackage{graphicx}
\usepackage{booktabs}
\def\red#1{\textcolor[rgb]{1,0,0}{#1}}
\def\blue#1{\textcolor[rgb]{0,0,1}{#1}}
\usepackage{bigstrut} 
\usepackage{makecell} 
\usepackage{multirow}
\usepackage[accsupp]{axessibility}  

\usepackage{hyperref}

\usepackage{orcidlink}

\begin{document}

\title{MutDet: Mutually Optimizing Pre-training for Remote Sensing Object Detection} 
\titlerunning{MutDet}

\author{Ziyue Huang\inst{1} \and 
Yongchao Feng\inst{1}
Qingjie Liu\inst{1,2*} \and
Yunhong Wang\inst{1,2}}

\authorrunning{Z.~Huang et al.}


\institute{State Key Laboratory of Virtual Reality Technology and Systems, Beihang University \and Hangzhou Innovation Institute, Beihang University \\
\email{\{ziyuehuang, fengyongchao, qingjie.liu, yhwang\}@buaa.edu.cn}
}

\maketitle

\begin{abstract}

Detection pre-training methods for the DETR series detector have been extensively studied in natural scenes, \textit{e.g.}, DETReg. 
However, the detection pre-training remains unexplored in remote sensing scenes. 
In existing pre-training methods, alignment between object embeddings extracted from a pre-trained backbone and detector features is significant. 
However, due to differences in feature extraction methods, a pronounced feature discrepancy still exists and hinders the pre-training performance. 
The remote sensing images with complex environments and more densely distributed objects exacerbate the discrepancy. 
In this work, we propose a novel \textbf{Mut}ually optimizing pre-training framework for remote sensing object \textbf{Det}ection, dubbed as \textbf{MutDet}. 
In MutDet, we propose a systemic solution against this challenge. 
Firstly, we propose a mutual enhancement module, which fuses the object embeddings and detector features bidirectionally in the last encoder layer, enhancing their information interaction.
Secondly, contrastive alignment loss is employed to guide this alignment process softly and simultaneously enhances detector features' discriminativity. 
Finally, we design an auxiliary siamese head to mitigate the task gap arising from the introduction of enhancement module. 
Comprehensive experiments on various settings show new state-of-the-art transfer performance. 
The improvement is particularly pronounced when data quantity is limited. When using 10 \% of the DIOR-R data, MutDet improves DetReg by 6.1\% in AP$_{50}$. 
\textit{Codes and models are available at: \href{https://github.com/floatingstarZ/MutDet}{https://github.com/floatingstarZ/MutDet}.}

\end{abstract}

\section{Introduction}
\label{sec:intro}

DETR-based methods \cite{ars, EMO2_DETR, rhino, o2detr} have recently been successfully applied to oriented object detection \cite{DOTA} in remote sensing images. 
However, DETR comes with training and optimization challenges, which need a large-scale training dataset due to the increased parameters in detection modules. 
In remote sensing images, the objects are densely distributed in the overhead view with arbitrary orientation, which requires more time and expert knowledge for annotation. 
The high annotation cost makes it difficult to obtain large-scale annotated datasets. 
We aim to address these challenges through detection pre-training \cite{soco, up_detr, detreg, aligndet, siamese_detr, presoco}, wherein detection modules are unsupervised pre-trained using generated pseudo-labels. 

\begin{figure}[tb]
  \centering
  \begin{subfigure}{0.45\linewidth}
  \includegraphics[height=1.9cm]{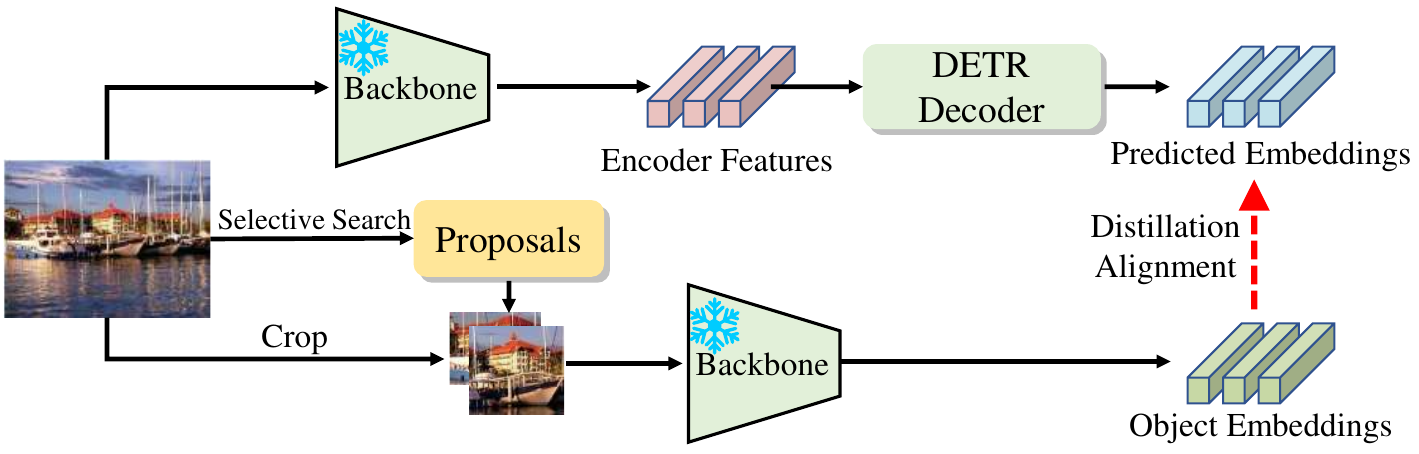}
    \caption{Predictive Approaches \cite{up_detr, detreg}}
    \label{fig:motivation_predict}
  \end{subfigure}
  \hfill
  \begin{subfigure}{0.45\linewidth}
  \includegraphics[height=1.9cm]{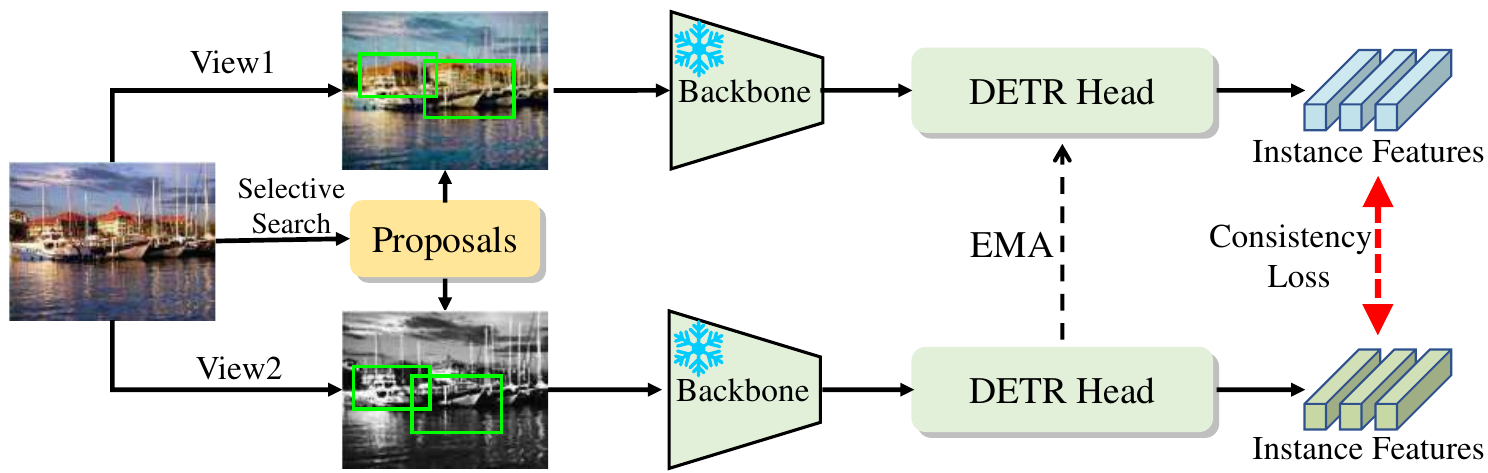}
    \caption{Self-supervised Learning \cite{aligndet, presoco}}
    \label{fig:motivation_self}
  \end{subfigure}
  \hfill
  \begin{subfigure}{0.45\linewidth}
  \includegraphics[height=1.9cm]{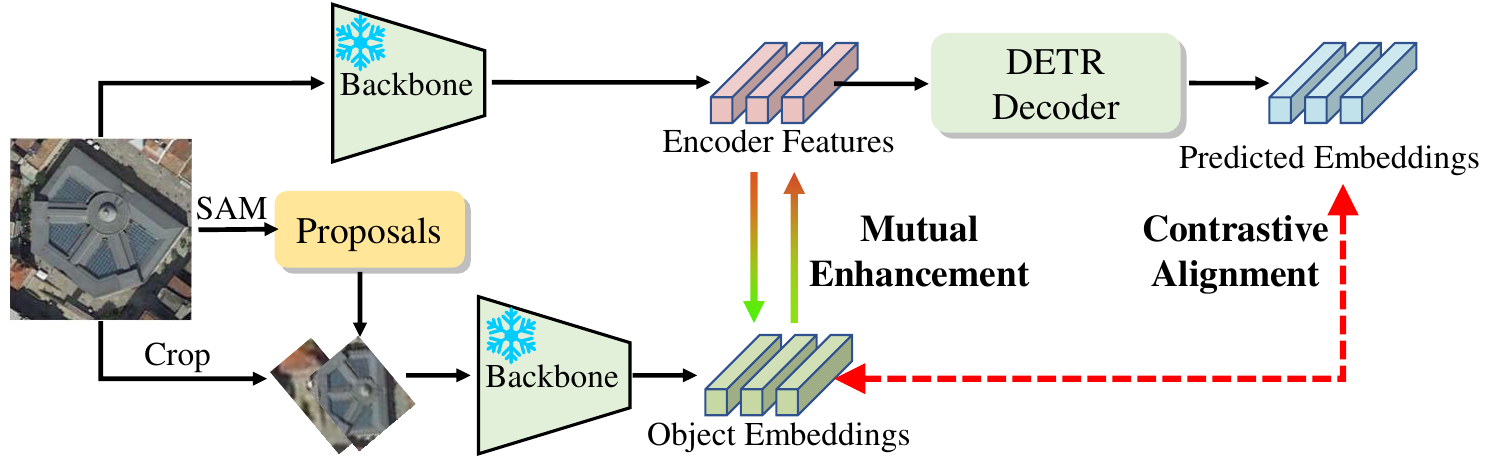}
    \caption{Mutually Optimizing (Ours)}
    \label{fig:motivation_mut}
  \end{subfigure}
    \hfill
  \begin{subfigure}{0.45\linewidth}
  \includegraphics[height=1.9cm, width=6.0cm]{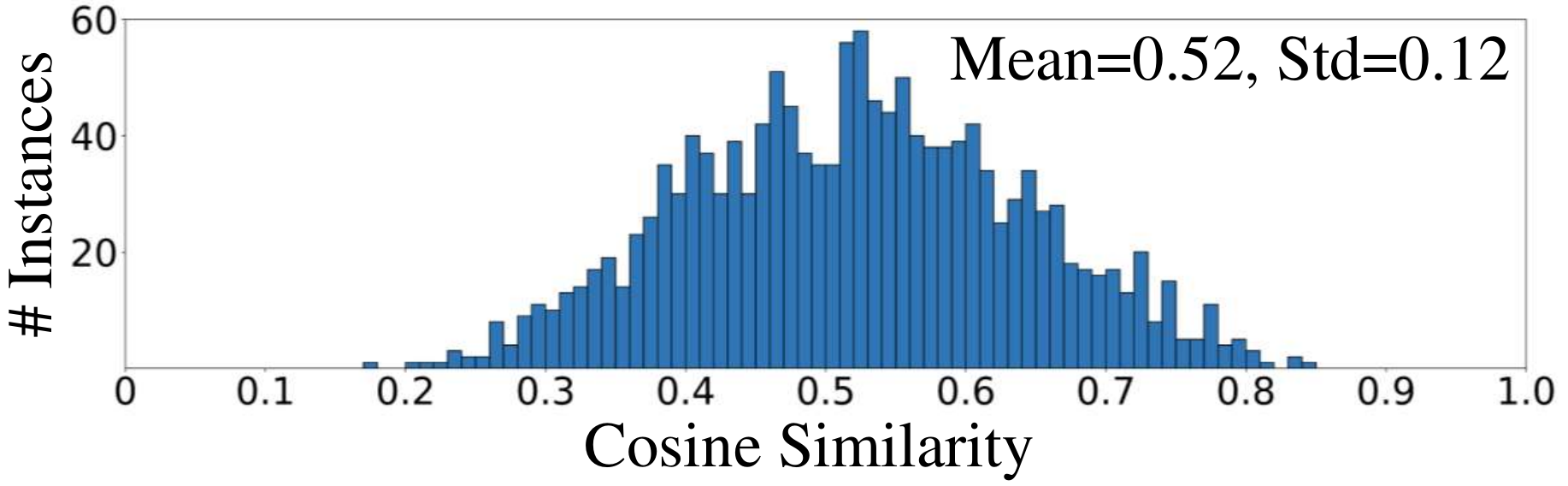}
    \caption{Feature Discripancy}
    \label{fig:motivation_feat}
  \end{subfigure}
  \caption{
  Motivation of our method. 
  (a) The predictive approaches \cite{detreg, up_detr} utilize the embedding alignment task to learn visual knowledge from the pre-trained backbone. 
  The feature discrepancy \cite{presoco} between object embeddings and detector features impedes the effectiveness of pre-training. 
  (b) Methods based on self-supervised learning \cite{aligndet, presoco} circumvent feature discrepancy but can not sufficiently leverage the knowledge from pre-trained backbone. 
  (c) Our approach employs contrastive alignment to achieve mutual learning between object embeddings and predictions, alleviating feature discrepancy. 
  Simultaneously, we enhance the learning of visual knowledge by deeply fusing object embeddings with encoder features. 
  (d) We use cosine similarity to measure the distance between object embeddings and predictions.  
  The detector fails to fit the object embeddings, indicating the feature discrepancy problem. 
  }
  \label{fig:motivation}
\end{figure}

Detection pre-training can broadly be categorized into predictive and self-supervised learning approaches \cite{huang2022survey}. 
Predictive approaches (Figure \ref{fig:motivation_predict}) such as UP-DETR \cite{up_detr} and DETReg \cite{detreg} achieve pre-training by making the detector to align the object embeddings of cropped images. 
The alignment is achieved by distillation-like alignment loss, enabling the model to learn fine-grained local features required for detection. 
However, there still exists a significant feature discrepancy \cite{presoco} (Figure \ref{fig:motivation_feat}) between the detector features and the object embeddings, hindering the pre-training performance. 
The feature discrepancy mainly arises from different feature extraction ways. 
The detector utilizes the image feature and a DETR decoder to predict embeddings, while object embeddings are extracted from cropped images through an entire backbone. 
The object embeddings contain deeper visual features unaffected by contextual interference. 
The complex and dense distribution of objects in remote sensing images also exacerbates the discrepancy. 
Self-supervised learning methods (Figure \ref{fig:motivation_self}), such as AlignDet \cite{aligndet} and PreSoco \cite{presoco}, constrain the consistency of instance features across different views to achieve self-training, which misses valuable visual knowledge inherent in pre-trained backbone. 

To address these limitations, we propose a \textbf{Mut}ually optimizing pre-training framework for remote sensing object \textbf{Det}ection, dubbed as \textbf{MutDet} (Figure \ref{fig:motivation_mut}). 
We introduce a mutual enhancement module to alleviate feature discrepancy. 
Concretely, we utilize bidirectional cross attention layers to deeply fuse the object embeddings and the encoder feature of the detector. 
The enhanced encoder feature is employed for subsequent training, prompting the DETR head to acquire visual knowledge from the pre-trained backbone more effectively. 
A contrastive alignment loss is employed to achieve collaborative optimization between object embeddings and predictions, enhancing detector features' discriminability. 
During the fine-tuning stage, since object embeddings are not accessible, the addition of mutual enhancement module will lead to a task gap \cite{up_detr} between the pre-training and fine-tuning. 
To address this issue, we design a calibration mechanism by adding an auxiliary siamese head. 

In summary, our contributions are listed as follows:
(1) We investigate pre-training methods for oriented object detection in remote sensing and propose a novel pre-training framework, \textit{i.e.}, MutDet. 
To the best of our knowledge, this is the first detection pre-training method in the field of remote sensing.  
(2) To mitigate the feature discrepancy issue, we present a mutually optimizing pre-training strategy, which includes the mutual enhancement module and contrastive learning. 
We introduce an auxiliary siamese head to address the task gap between pre-training and fine-tuning. 
(3) We compare MutDet with three detection pre-training methods on three datasets. 
Compared to DETReg \cite{detreg} baseline, MutDet improved by 2.8\% on DIOR-R, 1.9\% on DOTA-v1.0, and 2.99\% on OHD-SJTU-L, respectively. 
The improvement is more significant in conditions with limited data quantity or training time in fine-tuning. 
On DIOR-R, when using 10\% of the data, MutDet outperforms DETReg by 6.1\% in AP$_{50}$ and 4.5\% in AP$_{75}$; 
when using 1/3 of the training time (12 epochs), MutDet outperforms DETReg by 4.8\% in AP$_{50}$ and 3.9\% in AP$_{75}$.

\section{Related Work}
\label{sec:related}
\subsection{Oriented Object Detection}
In the past decade, remarkable progresses \cite{roi_trans, gwd, kfiou, yu2023phase, redet, xie2021oriented, ars} have been made in the field of oriented object detection in remote sensing images. 
RoI Transformer \cite{roi_trans} first adapts the two-stage detection framework to oriented detection task. 
GWD \cite{gwd} and several related works \cite{kfiou, yu2023phase,zhao2024abfl} optimize the localization loss with respect to rotation annotations. 
ReDet \cite{redet} proposes a rotation-equivariant network and a rotation-invariant feature extractor. 
Oriented R-CNN \cite{xie2021oriented} designs oriented region proposal network for efficient detection. 
Recently, DEtection TRansformer (DETR) \cite{detr, deform, dino} methods have been applied to this field. 
These methods \cite{o2detr, ao2detr, rhino, ars} introduce detection modules tailored for rotation object perception.
ARS-DETR \cite{ars}, built upon Deformable-DETR \cite{deform}, proposes an angle classification method and a rotated deformable attention module, effectively enhancing DETR’s high-precision detection performance.
Previous research focuses on improvement in model structure, with a limited exploration of training paradigms. 
Our work indicates that introducing detection pre-training before finetuning on downstream datasets can accelerate convergence and effectively improve detection performance. 

\subsection{Pre-training for detection}
Using detection pre-training with fine-grained pretext tasks has been proven to effectively enhance the fine-tuning performance in natural scenes, especially for DETR-based detectors \cite{up_detr, detreg, siamese_detr, aligndet, presoco}. 
Detection pre-training commonly utilizes unsupervised region proposal algorithms (\textit{e.g.} Selective Search \cite{selctive_search}) to generate object regions for learning object localization, with various approaches to learning object representations. 
SoCo \cite{soco} introduces instance-level multi-scale contrastive learning to train all modules of convolutional detectors from scratch. 
However, training all modules, as with self-supervised image representation learning \cite{byol}, requires expensive training resources. 
Subsequent research \cite{up_detr, detreg, aligndet} usually freezes the well-pre-trained backbone and trains only the detection-related modules, which reduces costs and preserves the generalization of the backbone. 
UP-DETR \cite{up_detr} and DETReg \cite{detreg} use pre-trained backbone to generate object embeddings and learn object representations through alignment with embeddings. 
ProSeCo \cite{presoco} points out that discrepancies between object embeddings and detector features may hinder pre-training performance and proposes a self-supervised pre-training method based on a student-teacher architecture. 
Recent research has employed a multi-view contrastive learning framework for detection pre-training \cite{aligndet, siamese_detr}.
Unfortunately, few works explored the detection pre-training in remote sensing images. 
In this paper, we design a novel detection pre-training framework for remote sensing scenes, in which 
we propose a systemic solution to address the feature discrepancy issue.
 
\section{Method}
\label{sec:method}
\subsection{Detector and Preparation}
\label{sec:method_pre}

\textbf{Detector}.
We build our methods upon ARS-DETR \cite{ars}, a strong DETR-detector for remote sensing images. 
ARS-DETR adopts a two-stage detection paradigm \cite{deform}, comprising a backbone, a transformer encoder, a transformer decoder containing 6 decoder layers, and multiple prediction heads. 
The encoder integrates multi-scale features from the backbone and predicts 300 rough proposals, which are further refined by the decoder to obtain detection results through prediction heads. 

\textbf{Preparation}.
\label{sec:Preparation}
We employ a pipeline similar to DETReg \cite{detreg} to generate the pseudo-labels containing boxes, classes, and object embeddings. 
The well-trained Segment Anything Model (SAM) \cite{sam} is utilized to generate the oriented bounding boxes.
Then, we crop the image patches according to the boxes and utilize a pre-trained backbone to extract patch features.  
Inspired by AlignDet \cite{aligndet}, we collect the features and apply principal component analysis to reduce the dimension, resulting in 256-dimensional object embeddings. 
Simultaneously, we employ k-means to cluster the normalized object embeddings into 256 classes. 
To reduce pre-training costs, we perform bounding box generation, object embedding extraction, and clustering in an offline manner, \textit{i.e.}, the generated pseudo-labels remain unchanged throughout the pre-training process.

\subsection{Overview}

\begin{figure}[t]
  \centering
  \includegraphics[height=6cm]{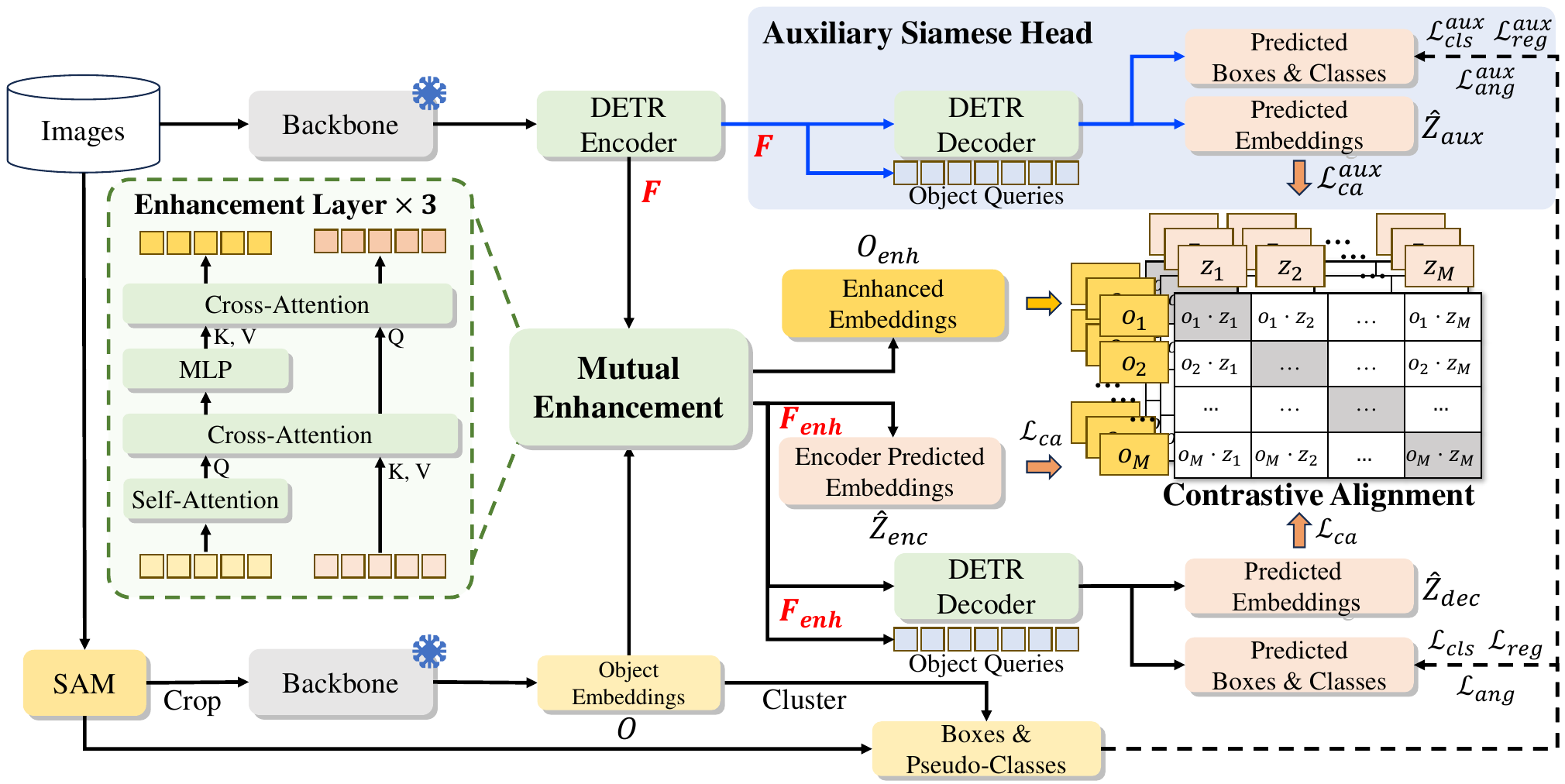}
  \caption{
  Overall architecture of the proposed MutDet. 
 MutDet optimizes DETReg \cite{detreg} and introduces SAM \cite{sam} to generate proposals. 
  It utilizes \textbf{mutual enhancement module} to cross-fuse the object embeddings and encoder features. 
  Then, it uses \textbf{contrastive alignment loss} to optimize the enhanced object embeddings and predicted embeddings mutually. 
  The \textbf{auxiliary siamese head} is proposed to alleviate the task gap between pre-training and fine-tuning, which shares parameters with the DETR decoder. 
  }
  \label{fig:method}
\end{figure}

The overview of our method is shown in Figure \ref{fig:method}. 
Given an input image $I$, we first obtain its pseudo-labels, including object embeddings $O$, boxes, and pseudo-classes as described in the Preparation. 
For the detector, the encoder features $F$ are produced by passing $I$ through a frozen backbone and the DETR encoder sequentially. 
Subsequently, $F$ along with $O$ is fed into \textbf{mutual enhancement module} (Sec. \ref{sec:mutual}), yielding enhanced encoder feature $F_{enh}$ and enhanced object embeddings $O_{enh}$. 
$F_{enh}$ is further utilized to obtain encoder predicted embeddings $\hat{Z}_{enc}$.
In conjunction with object queries, $F_{enh}$ is fed into the DETR decoder to predict the embeddings $\hat{Z}_{dec}$, boxes, and classes. 
To solve the task gap arising from the mutual enhancement module, we design an \textbf{auxiliary siamese head} (Sec. \ref{sec:siamese}), which shares parameters with the DETR decoder. 
This auxiliary head takes $F$ as input and predicts embeddings $\hat{Z}_{aux}$, boxes, and classes, similar to the DETR decoder. 
Finally, $\hat{Z}_{dec}$, $\hat{Z}_{enc}$, $\hat{Z}_{aux}$ and $O_{enh}$ are each paired as inputs of the \textbf{contrastive alignment loss} (Sec. \ref{sec:contrastive}) to calculate the alignment loss. 
Next, we describe each module in detail. 

\subsection{Mutual Enhancement Module}
\label{sec:mutual}
We utilize a mutual enhancement module to alleviate the feature discrepancy. 
Let $F \in \mathbb{R}^{K \times C}$ be the flattened multi-scale feature output by the DETR encoder, and $O \in \mathbb{R}^{M \times C}$ be the object embeddings, where $K$ denotes the number of sampling points, $M$ denotes the number of object embeddings, and $C=256$ denotes the feature dimension. 
The mutual enhancement module employs three enhancement layers to achieve bidirectional feature interaction. 
The $i$-th enhancement layer could be formulated as follows:
\begin{equation}
\label{eq:bi_attention1}
\begin{split}
&O' = \mathrm{LN}(\mathrm{MHSA}(O_i) + O_i), \quad\quad O''= \mathrm{LN}(\mathrm{MHCA}(O', F_i)+O') \\
&O_{i+1}=\mathrm{LN}(\mathrm{MLP}(O'')+O''), \quad F_{i+1}= \mathrm{LN}(\mathrm{MHCA}(F_i, O_{i+1})+F_i)
\end{split}
\end{equation}
where multi-head self-attention ($\mathrm{MHSA}$) layer, multi-head cross-attention ($\mathrm{MHCA}$) layer, layer normalization ($\mathrm{LN}$), and multi-layer perception ($\mathrm{MLP}$) are the typical modules in Transformer \cite{transformer}. 
$O_{i+1}$ and $F_{i+1}$ are the obtained fused object embeddings and encoder features of the $i+1$ layer, respectively.  
The final enhanced encoder feature $F_{enh}$ and enhanced object embeddings $O_{enh}$ will be employed for pre-training. 
$F_{enh}$ are fed into the decoder to predict embeddings, which are then aligned with $O_{enh}$, as shown in Figure \ref{fig:method}. 
In this process, we also introduce the DeNoising \cite{dino} strategy to obtain more accurate supervision signals. 


\begin{figure}[tb]
  \centering
    \setlength{\abovecaptionskip}{0.0cm}

  \begin{subfigure}{0.45\linewidth}
  \includegraphics[height=1.45cm]{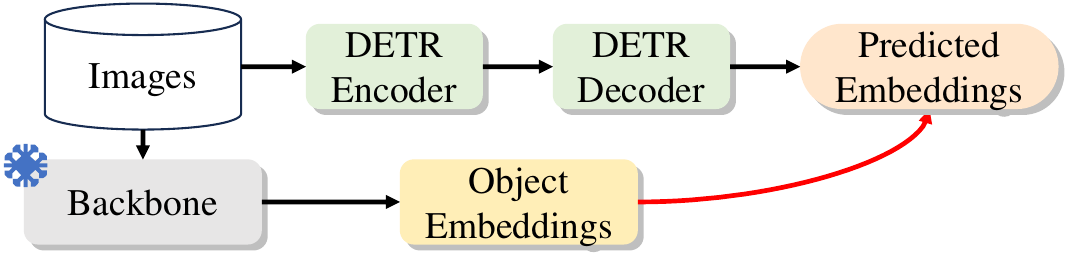}
    \caption{DETReg \cite{detreg}}
    \label{fig:supervision_detreg}
  \end{subfigure}
  \hfill
  \begin{subfigure}{0.45\linewidth}
  \includegraphics[height=1.45cm]{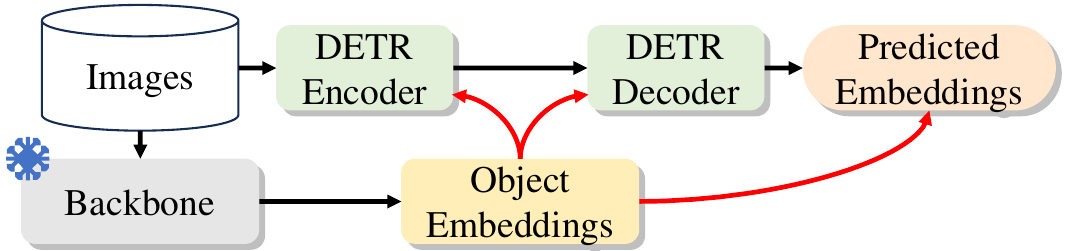}
    \caption{MutDet (Ours)}
    \label{fig:supervision_MutDet}
  \end{subfigure}
  \hfill
  \caption{
  A diagram illustrats the differences between the DETReg and MutDet with respect to supervisions. 
  DETReg is only supervised by object embeddings via one pathway in the last decoder layer. In contrast, our MutDet is supervised through multiple pathways. The \red{red} line represents the supervision signal.
  }
  \label{fig:supervision}
\end{figure}


From the perspective of backpropagation, DETReg \cite{detreg} can only receive supervised signals from object embeddings through predictions, which cannot be directly propagated to each module in the detector, as shown in Figure \ref{fig:supervision_detreg}. 
In MutDet, the mutual enhancement module integrates object embeddings with encoder features sufficiently during the forward propagation, allowing the supervised signals from object embeddings to influence modules in the detector through multiple pathways, as illustrated in Figure \ref{fig:supervision_MutDet}. Therefore, MutDet can receive diversified supervision from object embeddings and effectively learn visual knowledge from the pre-trained backbone.

\subsection{Alignment via Contrastive Learning}
\label{sec:contrastive}
We adopt contrastive alignment loss on enhanced embeddings $O_{enh}$ to accomplish alignment, driven by two main reasons. 
Firstly, instance-level contrastive alignment is equivalent to maximizing the mutual information between the distribution of object embeddings and predicted embeddings \cite{yang2022mutual}, thereby facilitating the learning of shared knowledge. 
Contrastive learning also prevents overfitting to semantics that might hinder generalization \cite{wang2022revisiting}. 
Secondly, since enhanced object embeddings become learnable, directly applying distillation loss will lead to feature collapse. 
In contrast, negative samples in contrastive learning solve this issue.  
The contrastive alignment loss between $M$ normalized object embeddings $O=\{\mathbf{o}_1, ..., \mathbf{o}_M\}$ and normalized predicted embeddings $Z=\{\mathbf{z}_1, ..., \mathbf{z}_M\}$ could be formulated as: 
\begin{small}
\begin{equation}
\mathcal{L}_{ca}(Z, O) = -\frac{2 \tau}{M}\sum_{i=1}^{M}{\left [
\log\frac{\exp(\mathbf{z}_i \cdot \mathbf{o}_i/\tau)}
{\sum_{k=1}^{M}{\exp(\mathbf{z}_i \cdot \mathbf{o}_k/\tau)}} + 
\log\frac{\exp(\mathbf{o}_i \cdot \mathbf{z}_i/\tau)}
{\sum_{k=1}^{M}{\exp(\mathbf{o}_i \cdot \mathbf{z}_k/\tau)}}
\right ]}
\label{eq:contrastive}
\end{equation}
\end{small}
where the temperature coefficient $\tau$ default set to 0.2 \cite{mocov3}, and we assume $O$ and $Z$ are one-to-one matched according to the index.

Here, we describe contrastive alignment loss in detection pre-training in detail. 
For an input image $I$ with $M$ annotations $\{\mathbf{b}_i, \mathbf{c}_i, \mathbf{a}_i, \mathbf{o}_i\}_{i=1}^{M}$, where $\mathbf{b}=(x, y, w, h)$ denotes the 4D bounding box, $\mathbf{c}\in \{1, 2, ..., 256\}$ denotes the category, $\mathbf{a}\in[-\pi/2, \pi/2)$ denotes the rotation angle, and $\mathbf{o}\in\mathbb{R}^{256}$ denotes the object embedding.
In MutDet, the DETR decoder outputs four predictions at each layer: 4D bounding boxes $\{\hat{\mathbf{b}}_i=(x_i, y_i, w_i, h_i)\}_{i=1}^{N}$, classification scores $\{\hat{\mathbf{p}}_i\in \mathbb{R}^{256}\}_{i=1}^{N}$, angle classification scores $\{\hat{\mathbf{a}}_i\in \mathbb{R}^{180}\}_{i=1}^{N}$ \cite{ars}, and predicted embeddings $\{\hat{\mathbf{z}}_i\in \mathbb{R}^{256}\}_{i=1}^{N}$, where $N$ denotes the number of object queries. 
Following DETReg \cite{detreg}, we apply contrastive alignment loss only to the final layer predictions. 
We perform Hungarian bipartite matching \cite{detr} to one-to-one match predictions $\{\hat{\mathbf{b}}_i, \hat{\mathbf{p}}_i, \hat{\mathbf{a}}_i\}_{i=1}^{N}$ and annotations $\{\mathbf{b}_i, \mathbf{c}_i, \mathbf{a}_i\}_{i=1}^{M}$ as the same in ARS-DETR \cite{ars}, resulting in the optimal permutation $\sigma$. 
Note that only the positive predicted embeddings $\hat{Z}_{dec}^{+}=\{\hat{\mathbf{z}}_{\sigma(i)}\}_{i=1}^{M}$ in matching are participate in the contrastive alignment loss. 
Besides, to further enhance supervision, we add extra contrastive alignment constraint on the enhanced encoder features $F_{enh}$. 
We feed $F_{enh}$ into the prediction heads of the encoder and select the top $N$ predictions according to the classification scores. 
Then, we also perform Hungarian bipartite matching to match the selected predictions and annotations $\{\mathbf{b}_i, \mathbf{c}_i, \mathbf{a}_i\}_{i=1}^{M}$ to get positive encoder predicted embeddings $\hat{Z}_{enc}^{+}$. 
The alignment loss $\mathcal{L}_{ca}^{det}$ of the detector is as follows:
\begin{equation}
\mathcal{L}_{ca}^{det} = \mathcal{L}_{ca}(\hat{Z}_{enc}^{+}, O_{enh}) + \mathcal{L}_{ca}(\hat{Z}_{dec}^{+}, O_{enh})
\label{eq:loss_ca}
\end{equation}

In addition to the embedding alignment task, detection pre-training includes localization and classification tasks \cite{aligndet, up_detr, detreg}. 
Here, the classification loss $\mathcal{L}_{cls}$ utilizes the focal loss \cite{focalloss} with the pseudo-labels obtained through clustering. 
The regression loss $\mathcal{L}_{reg}$ consists of the GIoU \cite{giou} loss and $L_1$ loss computed between the spatial coordinates (excluding angles) of predicted boxes and SAM-generated boxes. 
The angle loss $\mathcal{L}_{ang}$ uses the angle classification loss for oriented object detection \cite{ars}. 
Therefore, the overall loss of the detector is as follows:
\begin{equation}
\mathcal{L}_{det} = \mathcal{L}_{ca}^{det} + \mathcal{L}_{cls} + \mathcal{L}_{reg} + \mathcal{L}_{ang}
\label{eq:loss_all}
\end{equation}

\subsection{Auxiliary Siamese Head}
\label{sec:siamese}


Introducing the mutual enhancement module allows the detector to better fit the pre-training dataset. 
However, since object embeddings are not accessible during fine-tuning, feature enhancement cannot be performed. 
The enhancement module brings the task gap that affects the transferability of the pre-training.

Therefore, we consider a calibration mechanism to alleviate the gap. 
We aim that the original encoder feature $F$ to learn visual knowledge as effectively as the enhanced feature $F_{enh}$. 
We also expect the decoder to adapt to the distribution of $F$. 
Fortunately, self-distillation \cite{simkd, wang2023crosskd} can achieve this goal. 
Based on this clue, we attempt to introduce knowledge distillation for object detection \cite{zheng2022localization, yang2022prediction, chang2023detrdistill}. 
Three strategies are tested: encoder feature distillation, decoder cross distillation, and auxiliary siamese head, as shown in Table \ref{tab:calibration}. 
The encoder feature distillation is inspired by simple knowledge distillation \cite{simkd}, which employs $L_2$ feature distillation loss to align $F$ and $F_{enh}$. 
However, this approach fails to enable the decoder to adapt synchronously to the distribution of $F$. 
The decoder cross-distillation \cite{wang2023crosskd} feeds both $F$ and $F_{enh}$ into a shared decoder and uses distillation loss to align the outputs of the decoder. 
The distillation losses include knowledge distillation quality focal loss \cite{wang2023crosskd} for classification and $L_2$ loss for embedding alignment.
However, since $F_{enh}$ and the shared decoder continuously change during pre-training, obtaining accurate predicted embeddings and labels for effective distillation is challenging. 
Compared with the above two strategies, the simple siamese head yields the best performance.  
Different from the decoder cross-distillation, the auxiliary siamese head utilizes pseudo-labels to supervise the decoder output corresponding to $F$, as shown in Figure \ref{fig:method}. 
This approach allows the decoder to receive more precise and stable supervision signals. 
Furthermore, the shared decoder acts as an implicit constraint, guiding $F$ towards $F_{enh}$.

\begin{table}[t!]

  \centering

  \caption{The comparison of different calibration mechanisms.
  Pre-training is conducted using the DOTA-v1.0 dataset, and results on DIOR-R are reported, detailed settings are described to Sec. \ref{sec:exp_data}.
    Superscript denotes the improvement compared to without using calibration.  
  } 
    \setlength{\tabcolsep}{6pt}
    \begin{tabular}{c|cc|c|c}
    \hline
    \multirow{2}[4]{*}{} & \multicolumn{2}{c|}{Calibration Mechanism} & \multirow{2}[4]{*}{AP$_{50}$} & \multirow{2}[4]{*}{AP$_{75}$} \\
\cline{2-3}          & Encoder & Decoder &       &  \\
    \hline
    w/o calibration & -     & -     & 69.4  & 50.0  \\
    Encoder Feature Distillation & Distillation & -     & 69.6$^{+0.2}$  & 50.4$^{+0.4}$  \\
    Decoder Cross Distillation & - & Distillation & 69.9$^{+0.5}$  & 50.8$^{+0.8}$  \\
    Auxilary Siamese Head & - & Training & \textbf{70.7}$^{+1.3}$  & \textbf{51.2}$^{+1.2}$  \\
    \hline
    \end{tabular}%
  \label{tab:calibration}%
\end{table}%

We only apply contrastive alignment loss in Eq. \ref{eq:contrastive} to the shared decoder: 
\begin{equation}
\mathcal{L}_{ca}^{aux} = \mathcal{L}_{ca}(\hat{Z}_{aux}^{+}, O_{enh})
\label{eq:loss_ca_aux}
\end{equation}
where $\hat{Z}_{aux}^{+}$ denotes the positive predicted embeddings output by auxiliary siamese head. 
Meanwhile, we still utilize detection-related losses: $\mathcal{L}_{cls}^{aux}$, $\mathcal{L}_{reg}^{aux}$, and $\mathcal{L}_{ang}^{aux}$. And the loss of auxiliary siamese head is as follows:
\begin{equation}
\mathcal{L}_{det}^{aux} = \mathcal{L}_{ca}^{aux} + \mathcal{L}_{cls}^{aux} +  \mathcal{L}_{reg}^{aux} + \mathcal{L}_{ang}^{aux}
\label{eq:loss_aux}
\end{equation}
Ultimately, the overall loss for MutDet is:
\begin{equation}
\mathcal{L}_{mut} = \mathcal{L}_{det} + \mathcal{L}_{det}^{aux}
\label{eq:loss_total}
\end{equation}

\section{Experiments}
\label{sec:exp}
\subsection{Dataset and Implementation Details}
\label{sec:exp_data}

\textbf{Pre-training Datasets.} 
We only use the training and validation sets for pre-training to avoid data leakage. 
In this work, we choose DOTA-v1.0 \cite{DOTA} as the pre-training dataset and evaluate pre-training methods on multiple remote sensing detection datasets. 
Before the pre-training, images are divided into 800$\times$800 patches with an overlap of 200 pixels, resulting in 28,249 images. 

In addition, we perform large-scale pre-training by incorporating more high-quality data. 
We collect four large-scale remote sensing datasets and also divide the images into 800$\times$800 patches with an overlap of 200 pixels: DOTA \cite{DOTA} (28,249 images, 2,952,632 boxes), DIOR-R \cite{diorr} (11,725 images, 1,165,600 boxes), FAIR-1M-2.0 \cite{fair1m} (20,627 images, 2,519,081 boxes), and HRRSD \cite{hrssd} (29,916 images, 2,885,530 boxes), denoted as RSDet4. 
RSDet4 contains 90,518 images and nearly 10 million oriented bounding boxes, showcasing rich diversity and aligning well with remote sensing object detection tasks. 

\textbf{Pre-training Details.} 
We compare three detection pre-training methods: UP-DETR \cite{up_detr}, DETReg \cite{detreg}, and AlignDet \cite{aligndet}. 
Based on ARS-DETR, these methods are re-implemented in the MMRotate framework \cite{mmrotate} to adapt to oriented object detection tasks. 
We employ SAM to generate proposals to better handle directional, dense, and small objects in remote sensing.
Firstly, the SAM's automatic pipeline is employed to generate instance masks. 
We choose the ViT-H version of SAM \cite{sam}, utilizing a configuration with a $64 \times 64$ point grid and a Non-Maximum Suppression (NMS) threshold of 0.8, producing about 200 masks per image on average.
The instance masks are converted into oriented bounding boxes through the minimum bounding box algorithms. 
All boxes are utilized for pre-training to cover as many objects as possible. 
We find that the self-supervised pre-training (\textit{e.g.}, SwAV \cite{swav}) demonstrates inferior performance in remote sensing detection compared to supervised pre-training on ImageNet. 
Hence, we initialize the backbone with ImageNet pre-training. 
The backbone remains fixed during detection pre-training \cite{up_detr, detreg, aligndet}. 
We use AdamW optimizer \cite{adamw} with the initial learning rate of $1\times10^{-4}$ and train models on 4 NVIDIA GeForce RTX 3090 GPU with a total batch size of 8 for 36 epochs. 
We adopt a learning rate warm-up for 500 iterations, and the learning rate is reduced by a factor of 0.1 at the $32^{nd}$ epoch. 

\textbf{Fine-tuning Datasets.} 
Three datasets are selected to validate the transferability of pre-trained models. 
We use training set, validation set for fine-tuning and test set for evaluation. 
DIOR-R \cite{diorr} is an aerial image dataset with images from the DIOR \cite{DIOR} dataset, annotated with oriented bounding boxes. 
It contains 23,463 images and 192,518 instances in 20 common categories. 
All images in the dataset are in 800$\times$800. 
DOTA-v1.0 \cite{DOTA} contains 2,806 high-resolution remote sensing images with spatial resolutions ranging from 800 to 4,000 pixels, totaling 188,282 instances. 
Images in DOTA-v1.0 are divided into $1024 \times 1024$ patches with an overlap of 200 pixels without extra scaling. 
OHD-SJTU \cite{ohd} dataset consists of two subsets, namely OHD-SJTU-S and OHD-SJTU-L, containing 4,125 instances and 113,435 instances, respectively. 
Following ARS-DETR, we divide the images into 600$\times$600 patches with an overlap of 150 pixels and scale them to 800$\times$800. 

\textbf{Fine-tuning Details.} 
Models are initialized with detection pre-trained weights before fine-tuning, maintaining the same training hyper-parameters with pre-training, except that batch size is set to 4. 
In addition to the $3\times$ schedule for 36 epoch training, we test the $1\times$ schedule for 12 epoch training on OHD-SJTU, with the learning rate decay set at the $10^{th}$ epoch. 
The AP$_{50}$ and AP$_{75}$ under the DOTA evaluation protocol are reported. 
Furthermore, we separately report the AP$_{50}$ at 12, 24, and 36 epochs in the experiments.


\begin{table}[t]
  \centering
  \caption{Comparison results on DIOR-R \cite{diorr}. 
  All methods adopt ARS-DETR \cite{ars} as a detector and use ResNet-50 as the backbone. 
  Models are trained on the trainval set and evaluated on the test set. `-' indicates pre-training free, \textit{i.e.}, not using detection pre-training. 
  \textbf{\red{Red}}: optimal results. \textbf{\blue{Blue}}: sub-optimal results.}
    \resizebox{\linewidth}{!}{
    \begin{tabular}{c|cccccccccccccccccccc|cc}
    \hline
    Method & APL   & APO   & BF    & BC    & BR    & CH    & DAM   & ESA   & ETS   & GF    & GTF   & HA    & OP    & SH    & STA   & STO   & TC    & TS    & VE    & WM    &AP$_{50}$  &AP$_{75}$ \\
    \hline
    -  & 67.3  & 52.0  & 75.7  & 81.7  & 41.6  & 77.1  & 36.2  & 80.8  & 71.7  & 73.8  & 78.2  & 35.5  & 56.7  & 84.5  & 66.0  & 72.9  & 81.3  & 59.0  & 50.2  & 71.8  & 65.7  & 45.7  \\
    UP-DETR \cite{up_detr} & 68.1  & 54.7  & 77.6  & 81.9  & 42.5  & 78.5  & 33.9  & 82.9  & 73.8  & 78.5  & 78.3  & 43.4  & 55.4  & 85.6  & 69.1  & 73.3  & 81.0  & 61.0 & 50.5  & 71.7  & 67.1  & 48.4  \\
    AlignDet \cite{aligndet} & 67.6  & 53.4  & 73.4  & 81.2  & 40.4  & 77.1  & 38.4  & 81.1  & 71.6  & 73.4  & 78.1  & 35.8  & 54.7  & 84.6  & 69.7  & 73.2  & 80.7  & 59.7  & 49.6  & 71.8  & 65.8  & 45.8  \\
    DETReg \cite{detreg} & 69.0  & 55.5  & 75.7  & 82.4  & 45.4 & 78.6  & 35.6  & 81.7  & 72.4  & 78.8 & 78.2  & 46.1  & 57.7  & 87.6  & 71.5  & 76.0  & 81.3  & 60.6  & 52.9  & 72.1  & \textbf{\blue{67.9}}  & \textbf{\blue{49.1}}  \\
    MutDet (Ours) & 75.1 & 60.5 & 78.8 & 84.7 & 45.4 & 81.2 & 39.6 & 85.7 & 77.0 & 78.0  & 81.9 & 52.1 & 57.8 & 88.2 & 78.8 & 77.4 & 84.7 & 60.2  & 54.2 & 72.4 & \textbf{\red{70.7}} & \textbf{\red{51.2}} \\
    \hline
    \end{tabular}%
    }
  \label{tab:DIOR}%
\end{table}%

\begin{table}[t]
  \centering
  \caption{
  Comparison results on DOTA-v1.0 \cite{DOTA}. 
  All adopt ARS-DETR \cite{ars} as detector and use ResNet-50 as backbone. The results of the test set are reported. 
  `-' indicates pre-training free. 
  \textbf{\red{Red}}: optimal results. \textbf{\blue{Blue}}: sub-optimal results.}

    \resizebox{\linewidth}{!}{
    \begin{tabular}{c|ccccccccccccccc|cc}
    \toprule
    Method & PL    & BD    & BR    & GTF   & SV    & LV    & SH    & TC    & BC    & ST    & SBF   & RA    & HA    & SP    & HC    &AP$_{50}$  &AP$_{75}$  \\
    \midrule
    -  & 87.1  & 71.9  & 47.0  & 68.3  & 72.9  & 75.5  & 87.2  & 90.4  & 83.8 & 82.6  & 52.2  & 60.5  & 74.9  & 71.7  & 67.5 & \textbf{\blue{72.9}}  & 49.6  \\
    UP-DETR \cite{up_detr} & 80.1  & 77.0  & 48.9  & 70.2 & 74.6  & 76.3  & 87.7  & 90.6  & 78.4  & 82.9  & 53.3  & 66.0 & 75.3  & 70.9  & 59.1  & 72.7  & \textbf{\blue{50.0}}  \\
    AlignDet \cite{aligndet}  & 87.0  & 75.8  & 47.6  & 65.6  & 73.8  & 75.5  & 87.3  & 90.6  & 76.9  & 82.7  & 53.9  & 61.0  & 74.2  & 71.3  & 65.3  & 72.6  & 49.1  \\
    DETReg \cite{detreg} & 87.7 & 75.9  & 46.7  & 66.8  & 74.3  & 76.9  & 87.6  & 90.5  & 78.2  & 82.2  & 49.2  & 66.0 & 75.4  & 71.7  & 59.3  & 72.6  & 49.4  \\
    MutDet (Ours) & 87.3  & 78.7 & 51.3 & 68.5  & 78.9 & 81.6 & 88.1 & 90.7 & 79.9  & 83.7 & 58.0 & 61.8  & 76.5 & 72.1 & 60.8  & \textbf{\red{74.5}} & \textbf{\red{51.6}} \\
    \bottomrule
    \end{tabular}%
    }
  \label{tab:DOTA}%
\end{table}%

\begin{table}[t]
  \centering
  \setlength{\tabcolsep}{2pt}
  \caption{Comparison results on OHD-SJTU-S \cite{ohd} and OHD-SJTU-L \cite{ohd}. 
  Results for two training schedules (\textit{i.e.}, $1\times$ and $3\times$) at various IoU thresholds are reported.
  `-' indicates pre-training free. 
  \textbf{\red{Red}}: optimal results. \textbf{\blue{Blue}}: sub-optimal results.}
  \resizebox{\linewidth}{!}{
    \begin{tabular}{c|cccccc|cccccc}
    \hline
   \multirow{3}[6]{*}{Method} & \multicolumn{6}{c|}{OHD-SJTU-S \cite{ohd}}               & \multicolumn{6}{c}{OHD-SJTU-L \cite{ohd}} \\
\cline{2-13}    \multicolumn{1}{c|}{} & \multicolumn{3}{c}{1$\times$} & \multicolumn{3}{c|}{3$\times$} & \multicolumn{3}{c}{1$\times$} & \multicolumn{3}{c}{3$\times$} \\
\cline{2-13}    \multicolumn{1}{c|}{} &AP$_{50}$  &AP$_{75}$  & AP$_{50:95}$ &AP$_{50}$  &AP$_{75}$  & AP$_{50:95}$ &AP$_{50}$  & AP$_{75}$  & AP$_{50:95}$ & AP$_{50}$  &AP$_{75}$  & AP$_{50:95}$ \\
    \hline
    -  & 89.89  & 76.29  & 58.79  & 90.40  & 82.81  & 63.90  & 68.28  & 34.33  & 36.92  & \textbf{\blue{69.50}}  & 39.64  & 39.47  \\
    UP\text{-}DETR \cite{up_detr} & 90.27  & 82.85  & 67.79  & 89.73  & 82.97  & 66.72  & 70.69  & 42.15  & 40.88  & 69.48  & 43.23  & 40.44  \\
    AlignDet \cite{aligndet} & 89.80  & 70.71  & 58.06  & 90.40  & 80.84  & 64.09  & 68.39  & 37.81  & 37.82  & 69.02  & 40.43  & 39.52  \\
    DETReg \cite{detreg} & \textbf{\blue{90.56}}  & \textbf{\blue{83.23}}  & \textbf{\blue{68.40}}  & \textbf{\red{90.49}} & \textbf{\blue{83.31}}  & \textbf{\blue{68.83}}  & \textbf{\blue{70.87}}  & \textbf{\blue{44.79}}  & \textbf{\blue{41.87}}  & 68.61  & \textbf{\red{45.23}} & \textbf{\blue{41.03}}  \\
    MutDet (Ours) &  \textbf{\red{90.67}} &  \textbf{\red{88.09}} &  \textbf{\red{70.56}} & \textbf{\blue{90.41}}  & \textbf{\red{84.04}} &  \textbf{\red{69.82}} &  \textbf{\red{73.46}} &  \textbf{\red{45.06}} &  \textbf{\red{43.34}} &  \textbf{\red{71.60}} & \textbf{\blue{44.01}}  &  \textbf{\red{41.68}} \\
    \hline
    \end{tabular}%
    }
  \label{tab:OHD}%
\end{table}%

\subsection{Quantitative Results}

\textbf{Main Results.} 
We compare the performance of detection pre-training methods on three datasets: DIOR-R, DOTA-v1.0, and OHD-SJTU, as shown in Table \ref{tab:DIOR}, Table \ref{tab:DOTA}, and Table \ref{tab:OHD}, respectively. 
We also compare with the method that does not utilize detection pre-training, denoted as `pre-training free.' 
All pre-training methods use the training set of DOTA-v1.0 as the pre-training dataset and improve fine-tuning performance to some extent across all datasets. 
Our MutDet demonstrates consistent improvements over existing methods.
As shown in Table \ref{tab:DIOR}, MutDet achieves a significant improvement of 5.0\% in AP$_{50}$ compared to pre-training free. 
Moreover, MutDet improves by 2.8\% in AP$_{50}$ and by 2.1\% in AP$_{75}$ compared to the baseline DETReg. 
Similarly, our method achieves consistent improvement on DOTA-v1.0,  as shown in Table \ref{tab:DOTA}.  
Note that existing methods have all demonstrated negative impacts on fine-tuning, whereas MutDet still achieves a 1.6 \% improvement, demonstrating excellent stability. 
The pre-training failure may be attributed to using the same dataset for both pre-training and fine-tuning.
Consistency in data implies that it is challenging to acquire diverse object visual features distinct from the downstream task. 
Despite this limitation, MutDet still allows the detector to benefit from pre-training. 

According to the data scale, the OHD-SJTU dataset includes two subsets, OHD-SJTU-S and OHD-SJTU-L, containing 2 and 6 categories, respectively. 
Experiments are conducted on respective subsets with 1$\times$ and 3$\times$ training schedules. 
As shown in Table \ref{tab:OHD}, whether in 1$\times$ or 3$\times$ schedules, the improvement brought by detection pre-training on AP$_{75}$ is more pronounced than one on AP$_{50}$.
Under the 1$\times$ schedule on OHD-SJTU-S, compared to DetReg, MutDet improves by 0.11\%, 4.86\%, and 2.16\% in AP$_{50}$, AP$_{75}$, and AP$_{50:95}$, respectively. 
Under the 1$\times$ schedule on OHD-SJTU-L, MutDet improves by 2.59\%, 0.27\%, and 1.47\% in AP$_{50}$, AP$_{75}$, and AP$_{50:95}$, respectively. 
MutDet achieves either the optimal or sub-optimal results in all settings. 

\begin{table}[t!]
  \centering
      \setlength{\tabcolsep}{6pt}
  \caption{
  Object detection using k\% of the labeled data on DIOR-R. The models are trained on  k\%  trainval set and then evaluated on test set. 
  `-' indicates pre-training free. 
    Superscript denotes the improvement compared to pre-training free. 
  \textbf{\red{Red}}: optimal results. \textbf{\blue{Blue}}: sub-optimal results.
  }
    \resizebox{\linewidth}{!}{
    \begin{tabular}{c|cccc|cccc}
    \hline
    \multicolumn{1}{c|}{\multirow{2}[4]{*}{\newline{}Method}} & \multicolumn{4}{c|}{$AP_{50}$}     & \multicolumn{4}{c}{$AP_{75}$} \\
\cline{2-9}          & 10\%  & 25\%  & 50\%  & 100\% & 10\%  & 25\%  & 50\%  & 100\% \\
    \hline
    -  & 37.9  & 51.1  & 58.8  & 65.7  & 22.7  & 33.2  & 39.7  & 45.7  \\
    UP-DETR \cite{up_detr} & 49.9$^{+12.0}$  & 57.5$^{+6.4}$  & 62.0$^{+3.2}$  & 67.1$^{+1.4}$  & 35.0$^{+12.3}$  & 40.6$^{+7.6}$  & 44.4$^{+4.7}$  & 48.4$^{+2.7}$  \\
    AlignDet \cite{aligndet} & 37.9$^{+0.0}$  & 50.6$^{-0.5}$  & 58.3$^{-0.5}$  & 65.8$^{-0.1}$  & 23.2$^{+0.4}$  & 32.2$^{-1.0}$  & 39.3$^{-0.4}$  & 45.6$^{-0.1}$  \\
    DETReg \cite{detreg} & \textbf{\blue{50.8}}$^{+12.9}$  &  \textbf{\blue{58.4}}$^{+7.3}$  &  \textbf{\blue{63.1}}$^{+4.3}$  &  \textbf{\blue{67.9}}$^{+2.2}$  &  \textbf{\blue{35.8}}$^{+13.1}$  &  \textbf{\blue{41.5}}$^{+8.3}$  &  \textbf{\blue{45.4}}$^{+5.7}$  &  \textbf{\blue{49.1}}$^{+3.4}$  \\
    MutDet (Ours) & \textbf{\red{56.9}}$^{+19.0}$ & \textbf{\red{62.9}}$^{+11.8}$ & \textbf{\red{66.7}}$^{+7.9}$ & \textbf{\red{70.7}}$^{+5.0}$ & \textbf{\red{40.3}}$^{+17.6}$ & \textbf{\red{45.8}}$^{+12.6}$ & \textbf{\red{48.4}}$^{+8.7}$ & \textbf{\red{51.2}}$^{+5.5}$ \\
    \hline
    \end{tabular}%
    }
  \label{tab:low}%
\end{table}%

\textbf{Low Training Resources.} 
These experiments assess how detection pre-training methods perform when data quantity or training time is limited in the fine-tuning stage. 
Table \ref{tab:low} compares the performance of various methods when using k\% annotated data in fine-tuning. 
The less data available, the more improvement detection pre-training could be achieved compared with pre-training free, with our MutDet achieving the best in all settings. 
When using 10\% of the data, MutDet outperforms DETReg by 6.1\% in AP$_{50}$ and 4.5\% in AP$_{75}$. 
Our MutDet leverages only 50\% of the data to achieve comparable performance to pre-training free that uses 100\% of the data. 
These results indicate that MutDet can more effectively acquire visual knowledge from the pre-training backbone and dataset, improving the model's detection performance under data scarcity. 
Table \ref{tab:epoch} shows the detection performance at different epochs during fine-tuning. 
MutDet exhibits the most significant improvement at 12$^{nd}$ epoch, increasing by 11.5\% over pre-training free and by 4.8\% over DETReg in AP$_{50}$. 
At 24$^{th}$ epoch, MutDet still improves by 8.3\% over pre-training free and by 4.7\% over DETReg in AP$_{50}$. 

\begin{table}[t!]
\scriptsize
  \centering
    \setlength{\tabcolsep}{4pt}

  \caption{
  Comparison results at different epochs during training on DIOR-R. 
  All models are evaluated on test set at 12, 24, and 36 epochs. 
  `-' indicates pre-training free. 
      Superscript denotes the improvement compared to pre-training free. 
  \textbf{\red{Red}}: optimal results. 
  \textbf{\blue{Blue}}: sub-optimal results.
  }
    \resizebox{\linewidth}{!}{
    \begin{tabular}{c|cc|cc|cc}
    \hline
    \multirow{2}[2]{*}{Method} & \multicolumn{2}{c|}{12 Epoch} & \multicolumn{2}{c|}{24 Epoch} & \multicolumn{2}{c}{36 Epoch} \\
          &AP$_{50}$  &AP$_{75}$  &AP$_{50}$  & AP$_{75}$   &AP$_{50}$  & AP$_{75}$  \\
    \hline
    -  & 55.4  & 37.4  & 61.5  & 42.2  & 65.7  & 45.7 \\
    UP-DETR \cite{up_detr} & \textbf{\blue{62.5}}$^{+7.1}$ & \textbf{\blue{44.5}}$^{+7.1}$ & 64.7$^{+3.2}$ & 46.8$^{+4.6}$ & 67.1$^{+1.4}$ & 48.4$^{+2.7}$ \\
    AlignDet \cite{aligndet} & 54.3$^{-1.1}$ & 35.9$^{-1.5}$ & 60.7$^{-0.8}$ & 41.1$^{-1.1}$ & 65.8$^{+0.1}$ & 45.6$^{-0.1}$ \\
    DETReg \cite{detreg} & 62.1$^{+6.7}$ & 44.2$^{+6.8}$ & \textbf{\blue{65.1}}$^{+3.6}$ & \textbf{\blue{47.5}}$^{+5.3}$ & \textbf{\blue{67.9}}$^{+2.2}$ & \textbf{\blue{49.1}}$^{+3.4}$ \\
    MutDet (Ours) & \textbf{\red{66.9}}$^{+11.5}$ & \textbf{\red{48.1}}$^{+10.7}$ & \textbf{\red{69.8}}$^{+8.3}$ & \textbf{\red{49.9}}$^{+7.7}$ & \textbf{\red{70.7}}$^{+5.0}$ & \textbf{\red{51.2}}$^{+5.5}$ \\
    \hline
    \end{tabular}%
    }
  \label{tab:epoch}%
\end{table}%

\subsection{Ablation Study}
\begin{table}[t!]
\scriptsize
  \centering
  \caption{Effect of different components.  
  Models are evaluated on DIOR-R dataset, using DOTA-v1.0 as pre-training dataset, and ARS-DETR as detector. 
  The first line represents the DETReg baseline. 
    Superscript denotes the improvement compared to pre-training free. 
  }
\resizebox{\linewidth}{!}{

    \begin{tabular}{c|c|c|c|c|ccc}
    \hline
     Contrastive & Enhanced & Enhanced & Encoder & Siamase & \multicolumn{3}{c}{\#Epoch} \\
          Loss  & Embedding & Feature & Loss & Head  & 12    & 24    & 36 \\
    \hline
           &       &       &       &       & 62.1  & 65.1  & 67.9 \\
     \checkmark     &       &       &       &       & 62.6$^{+0.5}$  & 65.8$^{+0.7}$  & 68.5$^{+0.6}$ \\
    \checkmark     & \checkmark     &       &       &       & 62.8$^{+0.7}$  & 65.9$^{+0.8}$ & 68.7$^{+0.8}$ \\
     \checkmark     & \checkmark     & \checkmark     &       &       & 65.5$^{+3.4}$  & 67.0$^{+1.9}$    & 69.5$^{+1.6}$ \\
     \checkmark     & \checkmark     & \checkmark     & \checkmark     &       & 65.5$^{+3.4}$  & 67.4$^{+2.3}$  & 69.4$^{+1.5}$ \\
     \checkmark     & \checkmark     &       &       & \checkmark     & 64.4$^{+2.3}$ & 66.7$^{+1.6}$  & 69.1$^{+1.2}$ \\
    \checkmark     & \checkmark     &       & \checkmark     & \checkmark     & 66.3$^{+4.2}$  & 68.4$^{+3.3}$  & 69.9$^{+2.0}$ \\
     \checkmark     & \checkmark     & \checkmark     & \checkmark     & \checkmark     & \pmb{66.9}$^{+4.8}$ & \pmb{69.8}$^{+4.7}$  & \pmb{70.7}$^{+2.8}$ \\
    \hline
    \end{tabular}%
    }
  \label{tab:ablation}%
\end{table}%

\textbf{Effectiveness of individual component.} 
MutDet incorporates three designs: mutual enhancement module (Sec. \ref{sec:mutual}), contrastive alignment loss (Sec. \ref{sec:contrastive}), and auxiliary siamese head (Sec. \ref{sec:siamese}). 
We analyze the impact of each component in these three designs on the fine-tuning performance as shown in Table \ref{tab:ablation}. 
Upon the original DETReg, we introduce contrastive alignment loss and the enhanced embeddings from the mutual enhancement module. 
These two components improve the baseline performance to varying degrees. 
Subsequently, we feed the enhanced feature to DETR decoder, which significantly enhances fine-tuning performance, mainly improving by 3.4\% at the 12$^{nd}$ epoch in AP$_{50}$. 
Next, we apply additional contrastive alignment loss to the encoder. 
Although it does not directly improve performance, it synergizes with the subsequent auxiliary siamese head to further improve fine-tuning performance. 
Table \ref{tab:ablation} reveals that siamese head also contributes to the improvement, possibly due to introducing multiple supervision \cite{group}. 
Ultimately, the combination of all designs in MutDet achieves optimal performance.

\begin{table}[t!]
  \centering
      \setlength{\tabcolsep}{10pt}

  \caption{Comparison results with different backbones and pre-training datasets. 
  The performances on the test sets are reported. 
  Swin-Tiny denotes the smallest version of the Swin Transformer \cite{swin}. 
  RsDet4 is the collected large-scale pre-training dataset, including DOTA, DIOR, FAIR-1M-2.0, and HRRSD. 
  `-' indicates pre-training free. 
  \textbf{\red{Red}}: optimal results. \textbf{\blue{Blue}}: sub-optimal results.
  }
      \resizebox{1.0\linewidth}{!}{
    \begin{tabular}{c|c|c|c|ccc}
    \hline
    \multirow{2}[2]{*}{Method} & \multirow{2}[2]{*}{Backbone} & Pre-training & Fine-tuning & \multicolumn{3}{c}{\#Epoch} \\
          &       & Dataset & Dataset & 12    & 24    & 36 \\
    \hline
    -  & ResNet-50 & -     & \multirow{6}[2]{*}{DIOR} & 55.4  & 61.5  & 65.7  \\
    DETReg \cite{detreg} & ResNet-50 & DOTA-v1.0 &       & 62.1  & 65.1  & 67.9  \\
    MutDet (Ours) & ResNet-50 & DOTA-v1.0 &       & 66.9  & 69.8  & 70.7  \\
    -  & Swin-Tiny & -     &       & 58.1  & 64.9  & 70.0  \\
    DETReg \cite{detreg}  & Swin-Tiny & RSDet4 &       & \textbf{\blue{68.7}}  & \textbf{\blue{70.5}}  & \textbf{\blue{73.2}}  \\
    MutDet (Ours) & Swin-Tiny & RSDet4 &       & \textbf{\red{71.5}}  & \textbf{\red{70.5}}  & \textbf{\red{73.7}}  \\
    \hline
    -  & ResNet-50 & -     & \multirow{6}[2]{*}{DOTA-v1.0} & 69.0  & 71.1  & 72.9  \\
    DETReg \cite{detreg}  & ResNet-50 & DOTA-v1.0 &       & 70.0  & 72.5  & 72.6  \\
    MutDet (Ours) & ResNet-50 & DOTA-v1.0 &       & 73.8  & 73.3  & 74.5  \\
    -  & Swin-Tiny & -     &       & 70.8  & 74.0  & 75.6  \\
    DETReg \cite{detreg}  & Swin-Tiny & RSDet4 &       & \textbf{\blue{74.7}}  & \textbf{\red{76.3}}  & \textbf{\blue{76.7}}  \\
    MutDet (Ours) & Swin-Tiny & RSDet4 &       & \textbf{\red{75.5}}  & \textbf{\blue{76.0}}  & \textbf{\red{76.9}}  \\
    \hline
    \end{tabular}%
    }
  \label{tab:large scale}%
\end{table}%

\textbf{Advanced Backbone and Larger Pre-training Dataset}. 
We replace the detector backbone ResNet-50 with Swin-Tiny \cite{swin} and use the large-scale RSDet4 dataset for pre-training. 
Table \ref{tab:large scale} demonstrates that pre-training methods remain effective even with a stronger backbone and larger pre-training dataset. 
Different detection pre-training methods achieve similar performance, \textit{e.g.}, at 36$^{th}$ epoch, MutDet outperforms DETReg by only 0.5\% on DIOR-R and by 0.2\% on DOTA-v1.0. 
The differences between pre-training methods are more pronounced when training time is limited, \textit{e.g.}, at 12$^{nd}$ epoch, MutDet surpasses DETReg by 2.8\% on DIOR-R and by 0.8\% on DOTA.

\textbf{Different Detection methods}. 
In addition to ARS-DETR, we also test three variations of Deformable-DETR (D-DETR): Rotated D-DETR \cite{deform}, CSL D-DETR \cite{ars}, and AR-CSL D-DETR \cite{ars}. 
We adopt the same training settings as ARS-DETR, and the results are shown in Table \ref{tab:detector}. 
Our MutDet outperforms other pre-training methods on all detection methods.
MutDet surpasses DETReg by 4.4\% on Rotated D-DETR, 2.4\% on CSL D-DETR, and 3.1\% on AR-CSL D-DETR in AP$_{50}$, demonstrates its adaptability across different detectors.

\begin{table}[t]
  \centering
  \scriptsize
        \setlength{\tabcolsep}{12.8pt}
  \caption{Comparison results with different detection methods on DIOR-R. 
  `-' indicates pre-training free. 
  \textbf{\red{Red}}: optimal results. \textbf{\blue{Blue}}: sub-optimal results.
  }
    \begin{tabular}{c|cc|cc|cc}
    \hline
    \multirow{2}[2]{*}{Methods} & \multicolumn{2}{c|}{Rotated D-DETR} & \multicolumn{2}{c|}{CSL D-DETR} & \multicolumn{2}{c}{AR-CSL D-DETR} \\
          & AP$_{50}$  & AP$_{75}$  & AP$_{50}$  & AP$_{75}$  & AP$_{50}$  & AP$_{75}$ \\
    \hline
    -     &  38.3     &   23.7   & 62.2      &  41.2    &   63.6     & 42.1 \\
    UP-DETR \cite{up_detr} & 38.8  & 24.5  & 64.7  & 45.1  & 65.0  & \textbf{\blue{45.6}}  \\
    AlignDet \cite{aligndet} & 40.3  & 24.9  & 61.4  & 41.3  & 63.1  & 41.7  \\
    DETReg \cite{detreg} & \textbf{\blue{41.4}}  & \textbf{\blue{26.8}}  & \textbf{\blue{65.6}}  & \textbf{\blue{45.6}}  & \textbf{\blue{65.2}}  & 44.9  \\
    MutDet (Ours) & \textbf{\red{45.8}}  & \textbf{\red{28.0}}  & \textbf{\red{68.0}}  & \textbf{\red{47.3}}  & \textbf{\red{68.3}}  & \textbf{\red{47.6}}  \\
    \hline
    \end{tabular}%
  \label{tab:detector}%
\end{table}%

\begin{figure}[tb]
  \centering
  \includegraphics[height=2.3cm]{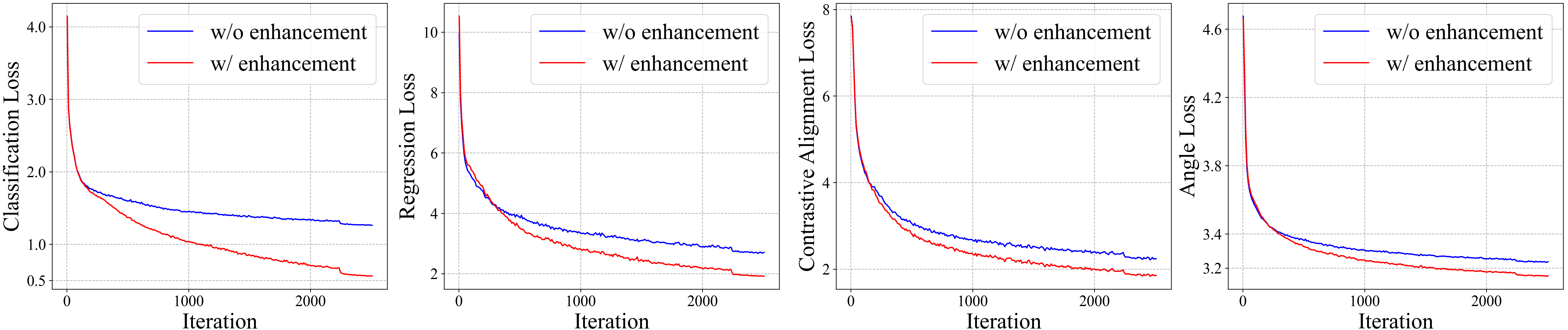}
  \caption{
  Loss curves during the pre-training on DOTA-v1.0 dataset. 
  The \red{red} curves are the losses when the mutual enhancement module is employed in pre-training, and the \blue{blue} curves are the losses when not employed. 
  Four losses are included: classification, regression, contrastive alignment, and angle prediction.
  }
  \label{fig:losses}
\end{figure}

\textbf{Effect of Mutual Enhancement Module. }
We compare the convergence of classification, regression, embedding alignment, and angle prediction tasks between using mutual enhancement module or not, as shown in Figure \ref{fig:losses}. 
After several training epochs, it is observed that the detector with the enhancement module converges more rapidly across all tasks. 
The phenomenon suggests that the detector can effectively acquire visual knowledge from the object embeddings through the enhancement module.

\section{Conclusion}
\label{sec:con}
In this work, we propose a novel pre-training framework for remote sensing detection. 
Our MutDet addresses the feature discrepancy issue in previous methods through mutual optimization, effectively improving the performance of downstream detection tasks. 
In MutDet, we introduce SAM to generate pseudo labels to enhance the recall of remote sensing objects and achieve rotation annotation. 
However, MutDet only unidirectionally utilizes SAM and fails to exploit its potential fully. 
Our future work will consider establishing a more effective correlation between the detector and the underlying visual model.
Our work lays a solid foundation for future pre-training research in remote sensing detection.

\section*{Acknowledgements}
This work was supported by the National Natural Science Foundation of China under Grant 62176017, and the Fundamental Research Funds for the Central Universities. 
%
%
\bibliographystyle{splncs04}
\bibliography{main}
\end{document}